\documentclass[sigconf, authorversion]{acmart}
\usepackage{epsfig}
\usepackage{graphicx}
\usepackage{amsmath}
\usepackage{booktabs}
\usepackage{array}
\usepackage{multirow}
\usepackage{color}
\usepackage{colortbl}
\usepackage{framed}
\usepackage{bm}
\usepackage{bbm}
\usepackage{xspace}
\usepackage{enumitem}
\usepackage{cite}
\usepackage{xparse}
\usepackage{xcolor}
\usepackage{algorithm}
\usepackage{lipsum}
\usepackage{comment}
\usepackage{listings}
\usepackage{url}
\usepackage[symbol]{footmisc}
\usepackage{pifont}
\definecolor{Gray}{gray}{0.9}
\definecolor{LightCyan}{rgb}{0.88,0.95,1}
\usepackage{balance}

\def \ie {\emph{i.e.}}
\def \eg {\emph{e.g.}}
\def \etal {\emph{et al.}}

\newcommand{\tit}[1]{\smallbreak\noindent\textbf{#1.}}
\newcommand{\tinytit}[1]{\noindent\textbf{#1.}}

\newcommand{\cmark}{\ding{51}}%
\newcommand{\xmark}{\ding{55}}%

\newcommand{\ours}{LaDI-VTON\xspace}
\newcommand{\dataset}{Dress Code\xspace}
\newcommand{\datasetviton}{VITON-HD\xspace}

\AtBeginDocument{%
  }

\setcopyright{acmcopyright}
\copyrightyear{2023}
\acmYear{2023}
\acmConference[MM '23]{Proceedings of the 31st ACM International Conference
on Multimedia}{October 29-November 3, 2023}{Ottawa, ON, Canada}
\acmBooktitle{Proceedings of the 31st ACM International Conference on
Multimedia (MM '23), October 29-November 3, 2023, Ottawa, ON, Canada}
\acmDOI{10.1145/3581783.3612137}
\acmISBN{979-8-4007-0108-5/23/10}

\begin{document}
\sloppy
\title[LaDI-VTON: Latent Diffusion Textual-Inversion Enhanced Virtual Try-On]{LaDI-VTON:\\Latent Diffusion Textual-Inversion Enhanced Virtual Try-On}



\author{Davide Morelli}
\authornote{Both authors contributed equally to this research.}
\orcid{0000-0001-7918-6220}
\affiliation{%
  \institution{University of Modena and Reggio Emilia}
  \city{Modena}
  \country{Italy}
}
\email{davide.morelli@unimore.it}

\author{Alberto Baldrati}
\authornotemark[1]
\orcid{0000-0002-5012-5800}
\affiliation{%
  \institution{University of Florence}
  \city{Florence}
  \country{Italy}
}
\email{alberto.baldrati@unifi.it}

\author{Giuseppe Cartella}
\orcid{0000-0002-5590-3253}
\affiliation{%
  \institution{University of Modena and Reggio Emilia}
  \city{Modena}
  \country{Italy}
}
\email{giuseppe.cartella@unimore.it}

\author{Marcella Cornia}
\orcid{0000-0001-9640-9385}
\affiliation{%
  \institution{University of Modena and Reggio Emilia}
  \city{Modena}
  \country{Italy}
}
\email{marcella.cornia@unimore.it}

\author{Marco Bertini}
\orcid{0000-0002-1364-218X}
\affiliation{%
  \institution{University of Florence}
  \city{Florence}
  \country{Italy}
}
\email{marco.bertini@unifi.it}

\author{Rita Cucchiara}
\orcid{0000-0002-2239-283X}
\affiliation{%
  \institution{University of Modena and Reggio Emilia}
  \city{Modena}
  \country{Italy}
}
\email{rita.cucchiara@unimore.it}


\begin{abstract}
The rapidly evolving fields of e-commerce and metaverse continue to seek innovative approaches to enhance the consumer experience. At the same time, recent advancements in the development of diffusion models have enabled generative networks to create remarkably realistic images. In this context, image-based virtual try-on, which consists in generating a novel image of a target model wearing a given in-shop garment, has yet to capitalize on the potential of these powerful generative solutions. This work introduces LaDI-VTON, the first Latent Diffusion textual Inversion-enhanced model for the Virtual Try-ON task. The proposed architecture relies on a latent diffusion model extended with a novel additional autoencoder module that exploits learnable skip connections to enhance the generation process preserving the model's characteristics. To effectively maintain the texture and details of the in-shop garment, we propose a textual inversion component that can map the visual features of the garment to the CLIP token embedding space and thus generate a set of pseudo-word token embeddings capable of conditioning the generation process. Experimental results on Dress Code and VITON-HD datasets demonstrate that our approach outperforms the competitors by a consistent margin, achieving a significant milestone for the task. Source code and trained models are publicly available at: \url{https://github.com/miccunifi/ladi-vton}.
\end{abstract}

\keywords{Virtual Try-On, Latent Diffusion Models, Generative Architectures.}

\begin{teaserfigure}
\centering
\includegraphics[width=0.97\textwidth]{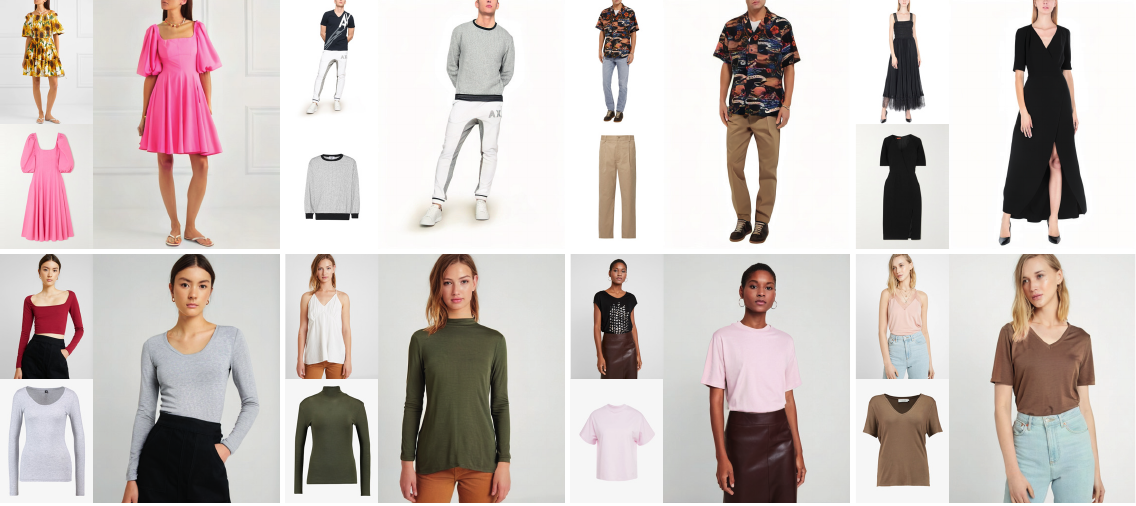}
\vspace{-.3cm}
\caption{Images generated by the proposed \ours model, given an input target model and a try-on clothing item from both Dress Code~\citep{morelli2022dresscode} (1st row) and VITON-HD~\citep{choi2021viton} (2nd row) datasets.}
\label{fig:teaser}
\end{teaserfigure}

\maketitle
\section{Introduction}
\label{sec:intro}
The disruptive success of e-commerce and online shopping is steadily demanding a more streamlined and enjoyable customer shopping experience, from personalized garment recommendation~\citep{hsiao2018creating,cucurull2019context,sarkar2023outfittransformer,de2023disentangling} to visual product search~\citep{hadi2015buy,liu2016deepfashion,wu2021fashion, baldrati2022conditioned, morelli2021fashionsearch}. Given the large availability of online images as accessories, garments, and other related products, Computer Vision and Multimedia research play a crucial role by offering valuable tools for a more personalized user experience. Among them, image-based virtual try-on has recently attracted significant interest in the research community with the introduction of several architectures~\citep{han2018viton,wang2018toward,morelli2022dresscode} that, given an image of a person and a garment taken from a catalog, allow to dress the person with the given try-on garment. 

The generation process carried out by current state-of-the-art methods for the task~\citep{morelli2022dresscode, lee2022high, ge2021parser, bai2022single} entirely relies on Generative Adversarial Networks (GANs)~\citep{goodfellow2014generative}. During the last years, a new family of generative architectures, namely diffusion models~\citep{sohl2015deep, ho2020denoising}, have shown superior image generation quality compared to GANs~\citep{dhariwal2021diffusion}, also with a more stable training procedure. However, considering the high computational demand typical of diffusion models, Rombach~\etal~\citep{rombach2022high} have recently tackled the problem by introducing a latent-based version that works in the latent space of a pre-trained autoencoder, thus finding the best trade-off between computational load and image quality.

Motivated by the tremendous success of these generative models, in this work, we introduce and explore for the first time an image-based virtual try-on method based on Latent Diffusion Models (LDMs)~\citep{rombach2022high}, demonstrating their successful possible applications in this field. We design a novel diffusion-based architecture conditioned on the target in-shop garment  
and human keypoints to keep the model's body pose unchanged. To preserve the target garment texture in the generation process, we propose to augment LDMs with a textual inversion network able to map the visual features of the in-shop garment to the CLIP textual token embedding space~\citep{Radford2021LearningTV}. We then condition the LDM generation through the cross-attention mechanism using the predicted tokens embeddings.

While LDMs can generate highly realistic images, one of their drawbacks is that they struggle when dealing with high-frequency details in the pixel space. This problem stems from the spatial compression performed by the autoencoder, which gives access to a lower-dimensional latent space where high-frequency details may not be accurately represented~\citep{rombach2022high}. In our setting, this can lead to details loss in the final generated images, especially when handling the model's hands, feet, and face. To address this issue, we introduce the Enhanced Mask-Aware Skip Connection (EMASC) module, a learnable skip connection that transfers the details from the encoding phase to the corresponding decoding one, improving the autoencoder reconstruction capabilities.

We extensively validate our architecture on two widely-used virtual try-on benchmarks (\ie, Dress Code~\citep{morelli2022dresscode} and VITON-HD~\citep{choi2021viton}), demonstrating superior quantitative and qualitative performance than state-of-the-art methods and showing that diffusion models applied to the virtual try-on field can achieve higher realism than GAN-based counterparts (Figure~\ref{fig:teaser}). 

\noindent \textbf{Contributions.} To sum up, our contributions are as follows:
\begin{itemize}[noitemsep,topsep=0pt]
    \item We employ LDMs to solve the task of image-based virtual try-on, an approach that, to the best of our knowledge, has never been previously explored in this field.
    \item To reduce the reconstruction error of LDMs, we enhance the autoencoder with learnable skip connections, enabling the preservation of details outside the inpainting region.
    \item Additionally, to increase detail retention of the generation process, we define a forward-only textual inversion module to further condition the model on the input try-on garment without losing texture information.
    \item Extensive experiments validate the effectiveness of each component of our architecture, which achieves state-of-the-art results on two widely used benchmarks for the task. We believe our results can highlight how virtual try-on can strongly benefit from using LDMs and serve as a starting point for future research in the field.
\end{itemize}

\section{Related Work}
\label{sec:related}
\tinytit{Image-Based Virtual Try-On}
Image-based virtual try-on~\citep{han2018viton, wang2018toward, issenhuth2020not, choi2021viton, fenocchi2022dual, morelli2022dresscode, bai2022single} aims to transfer a desired garment onto the corresponding region of a target subject while preserving human pose and identity. One of the pioneering works in this field is VITON~\citep{han2018viton}, a framework composed of an encoder-decoder generator that produces a coarse result further improved by a refinement network that exploits the warped clothing item obtained through a TPS transformation~\citep{duchon1977splines}. Some follow-up works have been oriented towards the enhancement of the warping module. Wang~\etal~\citep{wang2018toward} proposed a learnable TPS module to mitigate the problem of clothing details preservation, which has subsequently been improved either by combining TPS with affine transformations~\citep{fincato2021viton,li2021toward} or taking into account generated semantic layouts~\citep{yang2020towards} and body information~\citep{fele2022c}.

Another research line focuses on the generation phase and refinement of the result~\citep{issenhuth2020not,ge2021parser,morelli2022dresscode}.
Issenhuth~\etal~\citep{issenhuth2020not}, for example, presented a distillation-based teacher-student architecture that does not leverage a predicted semantic layout during the generation. This idea has further been explored in~\citep{ge2021parser} with the introduction of an additional tutor knowledge module to improve the generation quality. Differently, Morelli~\etal~\citep{morelli2022dresscode} focused on the semantics of the generated results and proposed a semantic-aware discriminator working at the pixel level instead of the image or patch level. Lee~\etal~\citep{lee2022high} solved the misalignment problem by designing a unified pipeline that combines the warping and segmentation stages to achieve better high-resolution results.

A common aspect linking all current methods is that the generation phase relies on GANs~\citep{goodfellow2014generative}. Driven by the enormous success of diffusion models~\citep{ho2020denoising} in different fields, we are the first, to the best of our knowledge, to propose an image-based virtual try-on architecture entirely relying on the aforesaid generative models.

\tit{Diffusion Models}
A fundamental line of research in the image synthesis field is the one marked by diffusion models~\citep{sohl2015deep, ho2020denoising, nichol2021improved, song2021denoising, ho2021classifier, dhariwal2021diffusion}. Inspired by non-equilibrium statistical physics, Sohl-Dickstein~\etal~\citep{sohl2015deep} defined a tractable generative model of data distribution by iteratively destroying the data structure through a forward diffusion process and then reconstructing with a learned reverse diffusion process. Some years later, Ho~\etal~\citep{ho2020denoising} successfully demonstrated that this process is applicable to generate high-quality images.
Nichol~\etal~\citep{nichol2021improved} further improved the work presented in~\citep{ho2020denoising} by learning the variance parameter of the reverse diffusion process and generating the output with fewer forward passes without sacrificing sample quality. While these methods work in the pixel space, Rombach~\etal~\citep{rombach2022high} proposed a variant working in the latent space of a pre-trained autoencoder, enabling higher computational efficiency.

\begin{figure*}[t]
  \centering
  \includegraphics[width=0.97\linewidth]{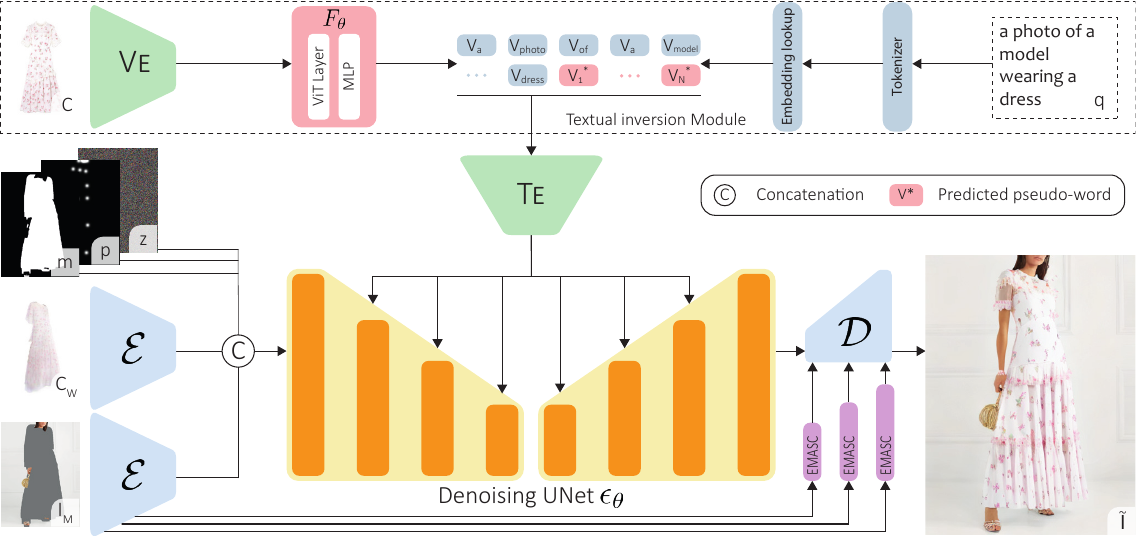}
  \vspace{-0.2cm}
  \caption{Overview of the proposed \ours model.
  On the top, the textual inversion module generates a representation of the in-shop garment. This information conditions the Stable Diffusion model along with other convolutional inputs.
  The decoder $\mathcal{D}$ is enriched with the Enhanced Mask-Aware Skip Connection (EMASC) modules to reduce the reconstruction error, improving the high-frequency details in the final image.
  }
  \label{fig:model}
  \vspace{-0.25cm}
\end{figure*}

The impact of diffusion models has rapidly become disruptive in diverse tasks such as text-to-image synthesis~\citep{ramesh2022hierarchical, sahariaphotorealistic, nichol2022glide, gu2022vector}, image-to-image translation~\citep{wang2022pretraining, saharia2022palette, zhaoegsde}, image editing~\citep{meng2022sdedit, avrahami2022blended, yang2023paint}, and inpainting~\citep{nichol2022glide, lugmayr2022repaint}.
Strictly related to virtual try-on is the task of human image generation, where pose preservation is often a strict constraint. On this line, Jiang~\etal~\citep{jiang2022text2human} focused on synthesizing full-body images given human pose and textual descriptions of shapes and textures of clothes, generating the output via sampling from a learned texture-aware codebook. Bhunia~\etal~\citep{bhunia2022person} tackled the task of pose-guided human generation by developing a texture diffusion block based on cross attention and conditioned on multi-scale texture patterns from the encoded source image. 
Baldrati~\etal~\citep{baldrati2023multimodal}, instead, proposed to guide the generation process constraining a latent diffusion model with the model pose, the garment sketch, and a textual description of the garment itself.

\tit{Textual Inversion}
Textual inversion is a recent technique proposed in~\citep{gal2022textual} to learn a pseudo word in the embedding space of the text encoder starting from visual concepts. 
Following~\citep{gal2022textual}, several promising methods~\citep{han2023highly, ruiz2022dreambooth, daras2022multires, mokady2022null} have been designed to enable personalized image generation and editing. Ruiz~\etal~\citep{ruiz2022dreambooth} presented a fine-tuning technique to bind an identifier with a subject represented by a few images and adopted a class-specific prior preservation loss to mitigate language drift. Similarly, Kumari~\etal~\citep{kumari2022customdiffusion} proposed a different fine-tuning method to enable multi-concept composition and showed that updating only a small subset of model weights is sufficient to integrate new concepts.
On a different line, Han~\etal~\citep{han2023highly} decomposed the CLIP embedding space~\citep{Radford2021LearningTV} based on semantics and enabled image manipulation without requiring any additional fine-tuning.

\section{Proposed Method}
\label{sec:method}
While most of the existing virtual try-on approaches leverage generative adversarial networks~\citep{han2018viton,wang2018toward,issenhuth2020not,morelli2022dresscode}, we propose a novel solution based, for the first time, on Latent Diffusion Models (LDMs). In particular, our work employs the Stable Diffusion architecture~\citep{rombach2022high} as a starting point to perform the virtual try-on task. To augment the text-to-image model with try-on capabilities, we modify the architecture to take as input both the try-on garment and the pose information of the target model. In addition, to better preserve the input clothing item details, we propose to add a novel forward-only textual inversion technique during the generation process. Finally, we enhance the image reconstruction autoencoder of Stable Diffusion with masked skip connections, thus improving the quality of generated images and better preserving the fine-grained details of the original model image. Figure~\ref{fig:model} depicts an overview of the proposed model.

\subsection{Preliminaries}

\tinytit{Stable Diffusion} It consists of an autoencoder $\mathcal{A}$ with an encoder $\mathcal{E}$ and a decoder $\mathcal{D}$, a text time-conditional U-Net denoising model $\epsilon_{\theta}$, and a CLIP text encoder $T_E$, which takes text $Y$ as input. The encoder $\mathcal{E}$ compresses an image $I \in \mathbb{R}^{3 \times H \times W}$ into a lower-dimensional latent space in $\mathbb{R}^{4 \times h \times w}$, where $h=\frac{H}{8}$ and $w=\frac{W}{8}$, while the decoder $\mathcal{D}$ performs the inverse operation and decodes a latent variable into the pixel space. 
For clarity, we refer to the $\epsilon_{\theta}$ convolutional input as the spatial input $\gamma$ (\eg, $z_t$) since convolutions preserve the spatial structure, and to the attention conditioning input as $\psi$ (\eg, $[t, T_E(Y)]$).
The training of the denoising network $\epsilon_{\theta}$ is performed by minimizing the following loss function:
\begin{equation}
L = \mathbb{E}_{\mathcal{E}(I), Y, \epsilon \sim \mathcal{N}(0,1),t} \left[ \lVert \epsilon - \epsilon_{\theta}(\gamma,\psi) \rVert_2^2 \right],
\label{eq:diffusion_loss}
\end{equation}
where $t$ represents the diffusing time step, $\gamma = z_t$, $z_t$ is the encoded image $\mathcal{E}(I)$ where we stochastically add Gaussian noise $\epsilon \sim \mathcal{N}(0,1)$, and $\psi=\left[t;T_E(Y)\right]$.

We aim to generate a new image $\Tilde{I}$ that replaces a target garment in the model input image $I$ with an in-shop garment $C$ provided by the user while retaining the model's physical characteristics, pose, and identity. This task can be seen as a particular type of inpainting, specialized in replacing garment information in human-based images according to a target garment image provided by the user. For this reason, we use the Stable Diffusion inpainting pipeline as the starting point of our approach. It takes as spatial input $\gamma$ the channel-wise concatenation of an encoded masked image $\mathcal{E}({I_M})$, a resized binary inpainting mask $m \in \{0,1\}^{1 \times h \times w}$, and the denoising network input $z_t$. Specifically, $I_M$ is the model image $I$ masked according to the inpainting mask $M \in \{0,1\}^{1 \times H \times W}$, and the binary inpainting mask $m$ is the resized version according to the latent space spatial dimension of the original inpainting mask $M$. To summarize, the spatial input of the inpainting denoising network is $\gamma = [z_t; m; \mathcal{E}(I_M)] \in \mathbb{R}^{(4+1+4) \times h \times w}$.

\tit{CLIP} It is a vision-language model~\citep{Radford2021LearningTV} which aligns visual and textual inputs in a shared embedding space. In particular, CLIP consists of a visual encoder $V_{E}$ and a text encoder $T_{E}$ that extract feature representations $V_{E}(I) \in \mathbb{R}^{d}$ and $T_{E}(E_L(Y)) \in \mathbb{R}^{d}$ for an input image $I$ and its corresponding text caption $Y$, respectively. Here, $d$ is the size of the CLIP embedding space, and $E_L$ is the embedding lookup layer which maps each $Y$ tokenized word to the token embedding space $\mathcal{W}$. 

The proposed approach introduces a novel textual inversion technique to generate a representation of the in-shop garment $C$. We feed this representation to the CLIP text encoder and use it to condition the diffusion process. It consists in mapping the visual features of $C$ into a set of $N$ new token embeddings $V_n^* \in \mathcal{W}, n=\{1,\ldots, N\}$. 
Following the terminology introduced in~\citep{baldrati2023zeroshot}, we refer to these embeddings as Pseudo-word Tokens Embeddings (PTEs) since they do not correspond to any linguistically meaningful entity but rather are a representation of the in-shop garment visual features in the token embedding space $\mathcal{W}$.

\subsection{Textual-Inversion Enhanced Virtual Try-On}
To tackle the virtual try-on task, we propose injecting in the Stable Diffusion textual conditioning branch additional information from the target garment $C$ extracted through textual inversion. In particular, starting from the features of the in-shop garment $C$ extracted from the CLIP visual encoder, we learn a textual inversion adapter $F_{\theta}$ to predict a set of fine-grained PTEs describing the in-shop garment $C$ itself. These PTEs lie in the CLIP token embedding space $\mathcal{W}$ and thus can be used as an additional conditioning signal. 

We also propose to extend the Stable Diffusion inpainting pipeline to accept the model pose map $P \in \mathbb{R}^{18 \times H \times W}$, where each channel is associated with a human keypoint, and the warped in-shop garment $C_W \in \mathbb{R}^{3 \times H \times W}$, representing the target garment $C$ warped according to the model body pose. While the pose map $P$ enables the method to preserve the original human pose of the model $I$, the warped garment $C_W$ helps the generation process to properly fit the garment onto the model. 
\tit{Data Preparation}
The warped garment $C_W$ is obtained by training a module that warps the in-shop garment $C$ fitting the model body shape in $I$. We employ the geometric matching module proposed in~\citep{wang2018toward} and refine the results with a U-Net-based component~\citep{ronneberger2015u}. 
The virtual try-on task involves replacing one or more garments the target model is wearing. With this aim, we define the inpainting area determined by the mask $M$ to fully encompass the target garment. We adopt the method proposed in previous works such as~\citep{morelli2022dresscode, issenhuth2020not} to ensure the mask completely covers the target garment.

\tit{Textual Inversion}
Given the in-shop image $C$, the aim of the textual inversion adapter $F_{\theta}$ is to predict a set of pseudo-word token embeddings $\{V_1^*, \ldots, V_N^*\}$ able to well represent the image $C$ in the CLIP token embedding space $\mathcal{W}$. We then use the predicted PTEs to condition the Stable Diffusion denoising network $\epsilon_{\theta}$ and obtain the final image $\tilde{I}$ where the model in $I$ is wearing the garment in $C$. For clarity, we intend that a set of PTEs represent well a target image if a Stable Diffusion model conditioned on the concatenation of a generic prompt and the predicted pseudo-words can reconstruct the target image itself.

We first build a textual prompt $q$ that guides the diffusion process to perform the virtual try-on task, tokenize it and map each token into the token embedding space using the CLIP embedding lookup module, obtaining $V_q$. Then, we encode the image $C$ using the CLIP visual encoder $V_E$ and feed the features extracted from the last hidden layer to the textual inversion adapter $F_{\theta}$, which maps the input visual features to the CLIP token embedding space $\mathcal{W}$. We then concatenate the prompt embedding vectors with the predicted pseudo-word token embeddings as follows: 

\begin{equation}
\label{eq:y_cappello}
\hat{Y} = \text{Concat}(V_q, F_{\theta}(V_E(C))).
\end{equation}
We feed the embedded concatenation $\hat{Y}$ to the CLIP text encoder $T_E$ and use the output to condition the denoising network $\epsilon_{\theta}$ leveraging the existing Stable Diffusion textual cross-attention.

To train the textual inversion adapter $F_{\theta}$, we use the inpainting pipeline of the out-of-the-box Stable Diffusion model as $\epsilon_{\theta}$. Specifically, it takes as input the encoded masked target model $\mathcal{E}(I_M)$, the inpainting mask $M$, and the latent variable $z$.
When training the adapter $F_{\theta}$, we freeze all the other model parameters.

To the best of our knowledge, this study marks the first instance in which a textual inversion approach has been employed in the domain of virtual try-on. As shown in the experimental section, this innovative conditioning methodology can significantly strengthen the final results and contribute to preserving the details and texture of the original in-shop garment.
Note that our proposed approach differs from traditional textual inversion techniques~\citep{gal2022textual, ruiz2022dreambooth, kumari2022customdiffusion}. Rather than directly optimizing the pseudo-word token embeddings through iterative methods, in our solution, the adapter $F_{\theta}$ is trained to generate these embeddings in a single forward pass.

\tit{Diffusion Virtual Try-On Model}
To perform the complete virtual try-on task, we employ the additional inputs described above (\ie, textual-inverted information $\hat{Y}$ of the in-shop garment, the pose map $P$, and the garment fitted to the model body shape $C_W$) to condition the Stable Diffusion inpainting pipeline.  
In particular, we extend the spatial input $\gamma \in \mathbb{R}^{9 \times h \times w}$ of the denoising network $\epsilon_{\theta}$ concatenating it 
with the resized pose map $p \in \mathbb{R}^{18 \times h \times w}$ and the encoded warped garment $\mathcal{E}(C_W) \in \mathbb{R}^{4 \times h \times w}$. The final spatial input results in $\gamma=\left[z_t; m; \mathcal{E}(I_M); p; \mathcal{E}(C_W) \right] \in \mathbb{R}^{(9+18+4) \times h \times w}$.

To enrich the input capacity of the denoising network $\epsilon_{\theta}$ without needing to retrain it from scratch~\citep{baldrati2023multimodal, rombach2022high}, we propose to extend the kernel channels of the first convolutional layer by adding zero-initialized weights to match the new input channel dimension. In such a way, we can retain the knowledge embedded in the original denoising network while allowing the model to deal with the newly proposed inputs.
Since the warped garment $C_W$ is not always able to properly represent the contextualization of the in-shop garment with the target model information, we also modify the Stable Diffusion textual input by using $\hat{Y}$ obtained from the output of the trained textual inversion adapter $F_{\theta}$ as described in Eq.~\ref{eq:y_cappello}.

As in standard LDMs, we train the proposed denoising network to predict the noise stochastically added to an encoded input $z_t = \mathcal{E}(I)$. We specify the corresponding objective function as:
\begin{equation}
\label{eq:final_loss}
    L = \mathbb{E}_{\mathcal{E}(I), \hat{Y}, \epsilon \sim \mathcal{N}(0,1), t, \mathcal{E}(I_M), M, p, \mathcal{E}(C_W)} \left[ \lVert \epsilon - \epsilon_{\theta}(\gamma,\psi) \rVert_2^2 \right],
\end{equation}
where $\psi=\left[t; T_E(\hat{Y})\right]$.

\begin{figure}[t]
  \centering
  \includegraphics[width=\linewidth]{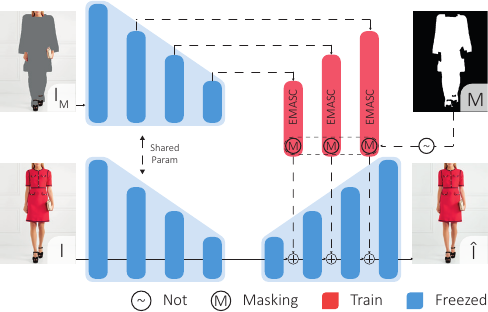}
  \vspace{-.5cm}
  \caption{Overview of the proposed autoencoder with Enhanced Mask-Aware Skip Connection (EMASC) modules.}
  \label{fig:EMASC}
  \vspace{-.3cm}
\end{figure}

\subsection{Enhanced Mask-Aware Skip Connections}
The autoencoder $\mathcal{A}$ of LDMs enables the denoising network $\epsilon_{\theta}$ to work within a latent space smaller than the pixel space. Compared to standard diffusion networks, this behavior is essential to reduce the parameters $\epsilon_{\theta}$ of the latent diffusion denoising network allowing it to reach the best trade-off between image quality and computational load~\citep{rombach2022high}. We remind that given an image $I \in \mathbb{R}^{3 \times H \times W}$, the Stable Diffusion encoder $\mathcal{E}$ compresses it in a latent space $Z \in \mathbb{R}^{4 \times \frac{H}{8} \times \frac{W}{8}}$, resulting in a total compression of $48 \times$.
However, this trade-off comes at a cost especially when dealing with human images and small high-frequency details such as hands, feet, and faces. We argue that the autoencoder reconstruction error partially depends on the data loss deriving from the latent space compression.

\begin{table*}[t]
    \caption{Quantitative results on the \dataset dataset~\citep{morelli2022dresscode}. The * marker indicates results reported in previous works, which may differ in terms of metric implementation. Best results are reported in \textbf{bold}.\vspace{-.2cm}}
    \label{tab:dresscode}
    \setlength{\tabcolsep}{.4em}
    \resizebox{0.98\linewidth}{!}{
    \begin{tabular}{lc cc c cc c cc c cccccc}
    \toprule
     & & \multicolumn{2}{c}{\textbf{Upper-body}} & & \multicolumn{2}{c}{\textbf{Lower-body}} & & \multicolumn{2}{c}{\textbf{Dresses}} & & \multicolumn{6}{c}{\textbf{All}} \\
    \cmidrule{3-4} \cmidrule{6-7} \cmidrule{9-10} \cmidrule{12-17}
    \textbf{Model} & & \textbf{FID$_\text{u}$} $\downarrow$ & \textbf{KID$_\text{u}$} $\downarrow$ & & \textbf{FID$_\text{u}$} $\downarrow$ & \textbf{KID$_\text{u}$} $\downarrow$ & & \textbf{FID$_\text{u}$} $\downarrow$ & \textbf{KID$_\text{u}$} $\downarrow$ & & \textbf{LPIPS} $\downarrow$ & \textbf{SSIM} $\uparrow$ & \textbf{FID$_\text{p}$} $\downarrow$ & \textbf{KID$_\text{p}$} $\downarrow$ & \textbf{FID$_\text{u}$} $\downarrow$ & \textbf{KID$_\text{u}$} $\downarrow$  \\
    \midrule
    PF-AFN*~\citep{ge2021parser} & & 14.32 & - & & 18.32 & - & & 13.59 & - & & - & - & - & - & - & - \\
    HR-VITON*~\citep{lee2022high} & & 16.86 & - & & 22.81 & - & & 16.12 & - & & - & - & - & - & - & - \\
    \midrule
    CP-VTON~\citep{wang2018toward} & & 48.31 & 35.25 & & 51.29 & 38.48 & & 25.94 & 15.81 & & 0.186 & 0.842 & 28.44 & 21.96 & 31.19 & 25.17 \\
    CP-VTON$^\text{\textdagger}$~\citep{wang2018toward} & & 22.18 & 12.09 & & 18.85 & 10.24 & & 21.83 & 12.31 & & 0.095 & 0.898 & 12.90 & 9.81 & 13.77 & 10.12 \\
    PSAD~\citep{morelli2022dresscode} & & 17.51 & 7.15 & & 19.68 & 8.90 & & 17.07 & 6.66 & & \textbf{0.058} & \textbf{0.918} & 8.01 & 4.90 & 10.61 & 6.17 \\
    \midrule
    \rowcolor{LightCyan}
    \textbf{\ours} & & \textbf{13.26} & \textbf{2.67} & & \textbf{14.80} & \textbf{3.13} & & \textbf{13.40} & \textbf{2.50} & & 0.064 & 0.906 & \textbf{4.14} & \textbf{1.21} & \textbf{6.48} & \textbf{2.20} \\
    \bottomrule
    \end{tabular}
    }
\vspace{-.2cm}
\end{table*}

To address the problem, we propose to extend the autoencoder architecture with an Enhanced Mask-Aware Skip Connection (EMASC) module whose aim is to learn to propagate relevant information from different layers of the encoder $\mathcal{E}$ to corresponding ones of the decoder $\mathcal{D}$. 
In particular, instead of skipping the information of the encoded image $I$ to reconstruct, we pass to the EMASC modules the intermediate features of the masked image $I_M$ encoding process, using the encoder $\mathcal{E}$. This procedure allows only the features not modified in the inpainting task to percolate, keeping the process cloth agnostic.
We implement EMASC employing additive non-linear learned skip connections in which we mask the output according to the inverted inpainting mask. Since the EMASC inputs are the intermediate features of the masked model $I_M$ encoding process, masking the EMASC output features helps avoid propagating the masked regions through the skip connections. Formally, the EMASC module is defined as follows:
\begin{equation}
\begin{split}
& EMASC_i = f(E_i) * NOT(m_i) \\
& D_i = D_{i-1} + EMASC_i
\end{split}
\label{eq:emasc}
\end{equation}
where $f$ is a learned non-linear function, $E_i$ is the $i$-th feature map coming from the encoder $\mathcal{E}$, $D_i$ is the corresponding $i$-th decoder feature map, and $m_i$ is obtained by resizing the mask $M$ according to the $E_i$ spatial dimension. An overview of the proposed autoencoder enhanced with EMASC modules is reported in Figure~\ref{fig:EMASC}.

Notice that the EMASC modules only depend on the Stable Diffusion denoising autoencoder, and once trained, they can be easily added to the standard Stable Diffusion pipeline in a plug-and-play manner without requiring additional training.
We show that this simple proposed modification can reduce the compression information loss in the inpainting task, resulting in better high-frequency human-related reconstructed details.

\section{Experimental Evaluation}
\label{sec:experiments}
\subsection{Datasets and Evaluation Metrics}
We perform experiments on two virtual try-on datasets, namely Dress Code~\citep{morelli2022dresscode} and VITON-HD~\citep{choi2021viton}, that feature high-resolution image pairs of in-shop garments and model images in both paired and unpaired settings. While in the paired setting the in-shop garment is the same as the model is wearing, in the unpaired one, a different garment is selected for the virtual try-on task. 

The Dress Code dataset~\citep{morelli2022dresscode} features over 53,000 image pairs of clothes and human models wearing them. The dataset includes high-resolution images (\ie, $1024 \times 768$) and garments belonging to different macro-categories, such as upper-body clothes, lower-body clothes, and dresses. In our experiments, we employ the original splits of the dataset where 5,400 image pairs (1,800 for each category) compose the test set and the rest the training one. The VITON-HD dataset~\citep{choi2021viton} instead comprises 13,679 image pairs, each composed of a frontal-view woman and an upper-body clothing item with a resolution equal to $1024 \times 768$. The dataset is divided into training and test sets of 11,647 and 2,032 pairs, respectively.

To quantitatively evaluate our model, we employ evaluation metrics to estimate the coherence and realism of the generation. In particular, we use the Learned Perceptual Image Patch Similarity (LPIPS)~\citep{zhang2018unreasonable} and the Structural Similarity (SSIM)~\citep{wang2004image} to evaluate the coherence of the generated image compared to the ground-truth. We compute these metrics on the paired setting of both datasets. To measure the realism, we instead employ the Fréchet Inception Distance~\citep{heusel2017gans} and the Kernel Inception Distance~\citep{binkowski2018demystifying} in both paired (\ie, FID$_\text{p}$ and KID$_\text{p}$) and unpaired (\ie, FID$_\text{u}$ and KID$_\text{u}$) settings. For the LPIPS and SSIM implementation, we use the torch-metrics Python package~\citep{detlefsen2022torchmetrics}, while for the FID and KID scores, we employ the implementation in~\citep{parmar2022aliased}.

\subsection{Implementation Details}
We first train the EMASC modules, the textual-inversion adapter, and the warping component. Then, we freeze the weights of all modules except for the textual inversion adapter and train the proposed enhanced Stable Diffusion pipeline\footnote{\url{https://huggingface.co/stabilityai/stable-diffusion-2-inpainting}}. In all our experiments, we generate images at $512\times 384$ resolution.
\begin{figure*}[t]
  \centering
  \includegraphics[width=0.98\linewidth]{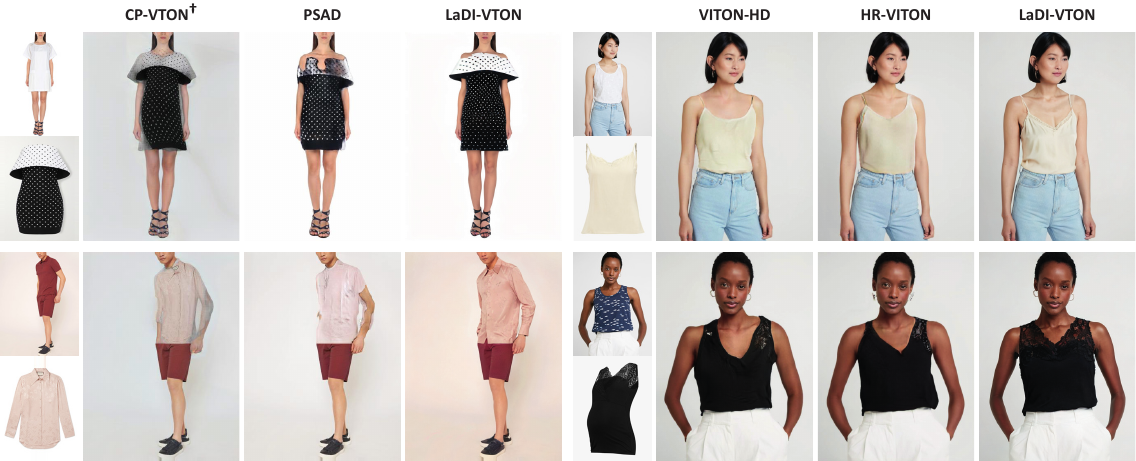}
  \vspace{-0.2cm}
  \caption{Qualitative results generated by \ours and competitors on Dress Code~\citep{morelli2022dresscode} (left) and VITON-HD~\citep{choi2021viton} (right).}
  \label{fig:method_qualitative}
  \vspace{-0.3cm}
\end{figure*}

\tit{Textual Inversion}
The textual inversion network $F_{\theta}$ consists of a single ViT layer followed by a multi-layer perception composed of three fully-connected layers separated by a GELU non-linearity~\citep{hendrycks2016gaussian} and a dropout layer~\citep{srivastava2014dropout}. We set the number of PTEs generated by $F_{\theta}$ to 16. We train $F_{\theta}$ for 200k steps, with batch size 16, learning rate 1e-5 with 500 warm-up steps using a linear schedule, AdamW~\citep{loshchilov2019decoupled} as optimizer with $\beta_1=0.9$, $\beta_2=0.999$, and weight decay equal to 1e-2. As the visual encoder $V_{E}$, we leverage the OpenCLIP ViT-H/14 model~\citep{wortsman2022robust} pre-trained on LAION-2B~\citep{schuhmann2022laionb}.

\tit{Diffusion Virtual Try-On Model}
We train the proposed virtual try-on pipeline for 200k iterations, with batch size 16 and the same optimizer and scheduling strategy used to train the textual inversion network. At training time, we randomly mask the text, the warped garment, and the pose map input with a probability of 0.2 for each condition. This allows the later use of the classifier-free guidance technique~\citep{ho2021classifier} at inference time. Following~\citep{avrahami2022spatext}, we use the fast variant of the multi-conditional classifier-free guidance, which allows computing the final result with a computational complexity independent from the amount of the input constraints.

\tit{Autoencoder with EMASC}
We apply the proposed EMASC modules to the variational autoencoder of the Stable Diffusion model. In particular, each EMASC module consists of two convolutional layers, where a SiLU non-linearity~\citep{elfwing2018sigmoid}  activates the first one. We apply the EMASC modules to the \texttt{conv\_in} layer output and the feature before the \texttt{down\_block} connecting each encoder layer to its corresponding decoder one. The convolutional layers have a kernel size of 3, padding of 1, and stride of 1. The first convolutional layer maintains the number of channels constant, while the second one adapts the channel axis dimension to the decoder features. Finally, we sum the EMASC output to the corresponding decoder features. We train the EMASC modules for 40k steps with batch size 16, learning rate equal to 1e-5, AdamW as optimizer with $\beta_1=0.9$, $\beta_2=0.999$, and weight decay 1e-2. Also, in this case, we perform 500 warm-up steps with a linear schedule. We employ a combination of the L1 and VGG~\citep{johnson2016perceptual} loss functions, scaling the perceptual VGG loss term by a factor of 0.5. In our setting, we found the VGG loss essential to avoid blurriness in the reconstructed images. During training, the encoder $\mathcal{E}$ and decoder $\mathcal{D}$ are frozen (see Figure~\ref{fig:EMASC}), and only the EMASC modules are learned.

\begin{table}[t]
    \caption{Quantitative results on the \datasetviton dataset~\citep{choi2021viton}. The * marker indicates results reported in previous works.\vspace{-0.2cm}}
    \label{tab:vitonhd_all}
    \setlength{\tabcolsep}{.3em}
    \resizebox{\linewidth}{!}{
    \begin{tabular}{lc cc cccc}
    \toprule
    \textbf{Model} & & \textbf{LPIPS} $\downarrow$ & \textbf{SSIM} $\uparrow$ & \textbf{FID$_\text{p}$} $\downarrow$ & \textbf{KID$_\text{p}$} $\downarrow$ & \textbf{FID$_\text{u}$} $\downarrow$ & \textbf{KID$_\text{u}$} $\downarrow$ \\
    \midrule
    CP-VTON*~\citep{wang2018toward} & & - & 0.791 & - & - & 30.25 & 40.12 \\
    ACGPN*~\citep{yang2020towards} & & - & 0.858 & - & - & 14.43 & 5.87 \\
    \midrule
    VITON-HD~\citep{choi2021viton} & & 0.116 & 0.863 & 11.01 & 3.71 & 12.96 & 4.09\\
    HR-VITON~\citep{lee2022high} & & 0.097 & \textbf{0.878} & 10.88 & 4.48 & 13.06 & 4.72\\
    \midrule
    \rowcolor{LightCyan}
    \textbf{\ours} & & \textbf{0.091} & 0.876 & \textbf{6.66} & \textbf{1.08} & \textbf{9.41} & \textbf{1.60} \\
    \bottomrule
    \end{tabular}
    }
\vspace{-.35cm}
\end{table}

\subsection{Experimental Results}
\tinytit{Comparison with State-of-the-Art Models}
We compare our method with several state-of-the-art competitors. For the Dress Code dataset, we compare our method with CP-VTON~\citep{wang2018toward} and PSAD~\citep{morelli2022dresscode}, retrained from scratch using the same image resolution of our model (\ie, $512\times384$) using the source codes when available or otherwise implementing them. Following~\citep{morelli2022dresscode}, we also include an improved version of CP-VTON (\ie, CP-VTON$^{\text{\textdagger}}$) where we add as additional input the masked image $I_M$. For the VITON-HD dataset, instead, we compare our model with VITON-HD~\citep{choi2021viton} and HR-VITON~\citep{lee2022high} using source codes and checkpoints released by the authors to extract the results. Given that some evaluation scores (\eg, LPIPS and FID) are very sensitive to different implementations, to ensure a fair comparison, we compute the quantitative results of these methods using the same metric implementation of our model. For completeness, we also include in the comparison some additional virtual try-on methods for which the results are from previous works and, therefore, may have been obtained using different evaluation source codes.

Table~\ref{tab:dresscode} reports the quantitative results on the Dress Code dataset. As can be seen, \ours achieves comparable results to PSAD~\citep{morelli2022dresscode} in terms of coherence with the inputs (\ie, LPIPS and SSIM), while significantly outperforming all competitors in terms of realism in both paired and unpaired settings. In particular,  on the Dress Code test set, our model reaches a FID score of 4.14 and 6.48 for the paired and unpaired settings, respectively. These results are considerably lower than the best-performing competitor (\ie, PSAD). In Table~\ref{tab:vitonhd_all}, we instead show the quantitative analysis of the VITON-HD dataset. Also, in this case, \ours surpasses all other competitors by a large margin in terms of FID and KID, demonstrating its effectiveness in this setting.

To qualitatively evaluate our results, we report in Figure~\ref{fig:method_qualitative} sample images generated by our model and by the competitors. Notably, our solution can generate high-realistic images and preserve the texture and details of the original in-shop garments, as well as the physical characteristics of target models.

\begin{table}[t]
\caption{User study results on the unpaired test set of both datasets. We report the percentage of times an image from \ours is preferred against a competitor.\vspace{-0.2cm}}
\label{tab:user_study}
\begin{center}
\footnotesize
\setlength{\tabcolsep}{.35em}
\resizebox{0.85\linewidth}{!}{
\begin{tabular}{l c c c c c c}
\toprule
\textbf{Dataset} & & \textbf{Model} & & \textbf{Realism} & & \textbf{Coherence} \\
\midrule
\multirow{3}{*}{Dress Code} & & CP-VTON~\citep{wang2018toward} & & 93.10 & & 89.68 \\
 & & $\text{CP-VTON}^{\text{\textdagger}}$~\citep{wang2018toward} & & 80.21 & & 75.69 \\
 & & PSAD~\citep{morelli2022dresscode} & & 74.14 & & 70.83 \\
\midrule
\multirow{2}{*}{VITON-HD} & & VITON-HD~\citep{choi2021viton} & & 79.19 & & 71.48 \\
& & HR-VITON~\citep{lee2022high} & & 77.95 & & 60.98 \\
\bottomrule
\end{tabular}
}
\end{center}
\vspace{-0.25cm}
\end{table}

\tit{Human Evaluation} To further evaluate the generation quality of our model, we conduct a user study to measure both the realism of generated images and their coherence with the inputs given to the virtual try-on model. Overall, we collect around 2,000 evaluations for each test, involving more than 50 unique users. In Table~\ref{tab:user_study}, we report the percentage of times in which an image generated by our model is preferred against a competitor. As can be seen, \ours is always selected more than 60\% of the time, further confirming the progress over previous methods.

\begin{table}[t]
    \caption{Quantitative results on the entire \dataset test set~\citep{morelli2022dresscode} using different model configurations.\vspace{-0.2cm}}  
    \label{tab:ablation_method}
    \setlength{\tabcolsep}{.28em}
    \resizebox{\linewidth}{!}{
    \begin{tabular}{lc cc cccc}
    \toprule

    \textbf{Model} & & \textbf{LPIPS} $\downarrow$ & \textbf{SSIM}  $\uparrow$ & \textbf{FID$_\text{p}$} $\downarrow$ & \textbf{KID$_\text{p}$} $\downarrow$ & \textbf{FID$_\text{u}$} $\downarrow$ & \textbf{KID$_\text{u}$} $\downarrow$ \\
    \midrule
    w/o text & & 0.071 & 0.902 & 4.99 & 1.61 & 8.50 & 3.70 \\
    w/ retrieved text & & 0.070 & 0.903 & 4.85 & 1.61 & 7.49 & 2.93\\
    w/ $F_{\theta}$ and standard SD & & 0.105 & 0.876 & 5.42 & 1.87 & 7.50 & 2.83\\
    w/o warped garment & & 0.068 & 0.904 & 4.50 & 1.44 & \textbf{6.30} & \textbf{1.99} \\
    \midrule
    \rowcolor{LightCyan}
    \textbf{\ours} & & \textbf{0.064} & \textbf{0.906} & \textbf{4.14} & \textbf{1.21} & 6.48 & 2.20\\
    \bottomrule
    \end{tabular}
    }
\vspace{-0.3cm}
\end{table}

\tit{Configuration Analysis}
In Table~\ref{tab:ablation_method}, we study the model performance by varying its configuration. We conduct this analysis on the Dress Code test set. In particular, the experiment in the first row replaces the Stable Diffusion textual input $\hat{Y}$ with an empty string. The one in the second row replaces the Stable Diffusion textual input $\hat{Y}$ with textual elements retrieved using the in-shop garment image $C$ as the query for a CLIP-based model~\citep{baldrati2023multimodal}. The results show that the proposed textual inversion adapter outperforms the other textual input alternatives. The third experiment regards the textual inversion adapter condition abilities, in particular, we can see that it is possible to obtain excellent results by using the proposed textual inversion adapter to condition an out-of-the-box Stable Diffusion model. Finally, we test the warped garment $C_W$ input in the overall pipeline by removing it. In this case, we can see that this additional input helps in the paired setting, but interestingly does not appreciably contribute to the unpaired one. 

\begin{table}[t]
    \caption{Quantitative analysis changing the number of predicted $V^*$. Results are reported on the Dress Code test set~\citep{morelli2022dresscode} using the out-of-the-box Stable Diffusion as backbone.\vspace{-0.2cm}}  
    \label{tab:ablation_vstar}
    \setlength{\tabcolsep}{.35em}
    \resizebox{0.9\linewidth}{!}{
    \begin{tabular}{cc cccccc}
    \toprule
    \textbf{\# $V^*$} & & \textbf{LPIPS} $\downarrow$ & \textbf{SSIM}  $\uparrow$ & \textbf{FID$_\text{p}$} $\downarrow$ & \textbf{KID$_\text{p}$} $\downarrow$ & \textbf{FID$_\text{u}$} $\downarrow$ & \textbf{KID$_\text{u}$} $\downarrow$ \\
    \midrule
    1 & & 0.115 & 0.867 & 6.14 & 2.24 & 8.19 & 3.14\\
    4 & & 0.108 & 0.873 & 5.87 & 2.15 & 8.17 & 3.10\\
    \rowcolor{LightCyan}
    16 & & 0.105 & 0.876 & 5.42 & 1.87 & 7.50 & 2.83\\
    32 & & 0.103 & 0.878 & 5.37 & 1.80 & 7.66 & 2.92\\
    \bottomrule
    \end{tabular}
    }
\vspace{-.25cm}
\end{table}

\begin{table}[t]
    \caption{Analysis on the effectiveness of the proposed Enhanced Mask Aware Skip Connection modules. Results are reported on Dress Code~\citep{morelli2022dresscode} and VITON-HD~\citep{choi2021viton}.\vspace{-0.2cm}}
    \label{tab:vae}
    \setlength{\tabcolsep}{.35em}
    \resizebox{0.9\linewidth}{!}{
    \begin{tabular}{lc ccc c cc}
    \toprule
     & & \textbf{Model} & \textbf{EMASC} & \textbf{Masked} & & \textbf{LPIPS} $\downarrow$ & \textbf{SSIM} $\uparrow$  \\
    \midrule
     & & SD VAE & None & - & & 0.0214 & 0.9538 \\
     & & SD VAE & Linear & \cmark & & 0.0196 & 0.9636 \\
     & & SD VAE & Non-Linear & \xmark & & 0.0183 & 0.9646 \\
    \rowcolor{LightCyan}
     \cellcolor{white} &  \cellcolor{white} & SD VAE & Non-Linear & \cmark & & \textbf{0.0181} & \textbf{0.9652} \\
    \cmidrule{3-8}
    & & \ours & None & - & & 0.0642 & 0.8985 \\
    \rowcolor{LightCyan}
     \cellcolor{white}\multirow{-7}{*}{{\rotatebox[origin=c]{90}{Dress Code}}} &  \cellcolor{white} & \ours & Non-Linear & \cmark & & \textbf{0.0640} & \textbf{0.9060} \\
    \midrule
    & & SD VAE & None & - & & 0.0260   & 0.9336 \\ 
     & & SD VAE & Linear & \cmark & & 0.0220 & 0.9545 \\
     & & SD VAE & Non-Linear & \xmark & & 0.0203 & 0.9560 \\
     \rowcolor{LightCyan}
    \cellcolor{white} &  \cellcolor{white} & SD VAE & Non-Linear & \cmark & & \textbf{0.0200} & \textbf{0.9561} \\
    \cmidrule{3-8}
    & & \ours & None & - & & 0.0960 & 0.8491 \\
    \rowcolor{LightCyan}
     \cellcolor{white}\multirow{-7}{*}{{\rotatebox[origin=c]{90}{VITON-HD}}} &  \cellcolor{white} & \ours & Non-Linear & \cmark & & \textbf{0.0907} & \textbf{0.8758} \\
    \bottomrule
    \end{tabular}
    }
\vspace{-0.3cm}
\end{table}

\tit{Analysis on $\mathbf{V^*}$}
In Table~\ref{tab:ablation_vstar}, we show the results of the out-of-the-box Stable Diffusion inpainting model conditioned using the textual inversion module when varying the number of PTEs generated by $F_{\theta}$. Overall, we obtain the best scores in terms of FID and KID on the unpaired setting using 16 pseudo-word tokens embeddings, while for the metrics on the paired setting employing 32 PTEs leads to slightly better results. Since increasing the number of PTEs can increase memory usage, the best trade-off between computational load and performance is reached when using 16 PTEs.

\tit{Effectiveness of EMASC modules}
We test the proposed EMASC modules on the paired settings of Dress Code and VITON-HD. In particular, we test the EMASC performance on both the autoencoder $\mathcal{A}$ for image reconstruction and the final model (\ie, \ours) for the complete virtual try-on task. In the first experiment, we simply encode and then decode the model image $I$ obtaining the reconstructed image $\hat{I}$ (\ie, $\hat{I} = \mathcal{D}(\mathcal{E}(I))$). In the second experiment, we compare the performance of our complete model with and without the EMASC modules. While for the first experiment, LPIPS and SSIM are computed by comparing the model image $I$ with its reconstruction $\hat{I}$, in the second experiment, we evaluate the metrics by comparing the model image $I$ with its reconstruction $\tilde{I}$, where we define $\tilde{I}$ as the output of the virtual try-on pipeline. Results reported in Table~\ref{tab:vae} show that the proposed method can enhance both the reconstruction capabilities of the Stable Diffusion autoencoder and the output performance of the final virtual try-on pipeline leading to better evaluation scores.

\begin{figure}[t]
  \centering
  \includegraphics[width=0.95\linewidth]{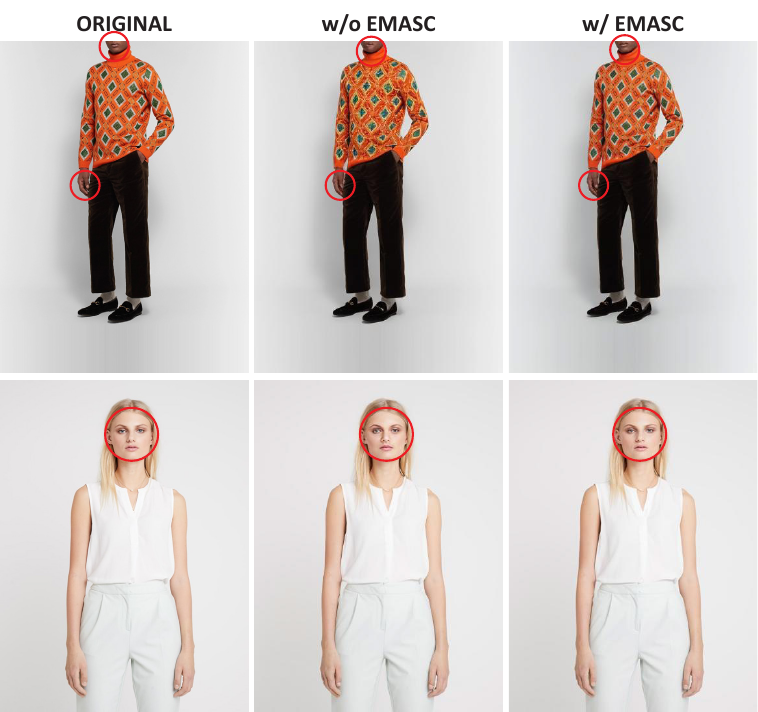}
  \vspace{-0.2cm}
  \caption{Image reconstruction results from the Stable Diffusion autoencoder with and without the EMASC modules.}
  \label{fig:vae_dresscode_qualitative}
  \vspace{-0.35cm}
\end{figure}

To better assess the contribution of the EMASC modules in the autoencoder analysis, we compare the proposed EMASC method with two of its variants. The first variant involves removing the feature masking after the final convolutional layer, while in the second variant, we use only one convolutional layer without any non-linear activation. 
We can notice that the masked non-linear EMASC modules achieve better results in all metrics on both datasets. In Figure~\ref{fig:vae_dresscode_qualitative}, we also show sample qualitative results of the Stable Diffusion autoencoder with and without EMASC modules. As it is possible to see, the proposed learnable mask-aware skip-connections reduce the reconstruction loss resulting in better faces, hands, and feet. Note that we achieve such results without retraining or fine-tuning the autoencoder.

\section{Conclusion}
\label{sec:conclusion}
In this work, we propose the first latent diffusion-based approach for virtual try-on. To increase the detail retention of the input in-shop garment, we exploit the textual inversion technique for the first time in this task, demonstrating its capability in conditioning the generation process. Moreover, we introduce the EMASC modules that can enhance the inpainting output image quality reducing the autoencoder compression loss of LDMs. This advancement notably improves the human perceived quality of high-frequency human body details such as hands, faces, and feet. Results show that the proposed \ours model outperforms by a large margin the competitors in terms of realism on both Dress Code and VITON-HD datasets, two widely used benchmarks for the task.

\begin{acks}
This work has partially been supported by the European Horizon 2020 Programme (grant number 101004545 - ReInHerit) and by the PRIN project ``CREATIVE: CRoss-modal understanding and gEnerATIon of Visual and tExtual content'' (CUP B87G22000460001), co-funded by the Italian Ministry of University.
\end{acks}

\appendix
\section{Clothes warping procedure}
To warp the in-shop garment $C$ to fit the model's body shape shown in $I$ and obtain the warped in-shop garment $C_W$, we exploit the geometric matching module proposed in~\citep{wang2018toward} and a U-Net based~\citep{ronneberger2015u} refinement component.

Specifically, the geometric matching module computes a correlation map between the encoded representations of the in-shop garment $C$ and a cloth-agnostic person representation composed of the pose map $P$ and the masked model image $I_M$. We obtain these encoded representations using two separate convolutional networks. Based on the computed correlation map, we predict the spatial transformation parameters $\theta$ of a thin-plate spline geometric transformation~\citep{duchon1977splines,rocco2017convolutional} represented by $\text{TPS}_{\theta}$. We use the $\theta$ parameters to compute the coarse warped garment $\hat{C}$ starting from the in-shop garment $C$ (\ie, $\hat{C} = \text{TPS}_{\theta}(C)$).

To further refine the result, we use a U-Net model that takes as input the concatenation of the coarse warped garment $\hat{C}$, the pose map $P$, and the masked model image $I_M$ and predicts the refined warped garment $C_W$ as follows:
\begin{equation}
C_W = \text{U-Net}(\hat{C}, P, I_M).
\end{equation}

\tit{Training details}
We first train the geometric matching module for 50 epochs with batch size 32 using the L1 loss function. Then, the U-Net refinement module is trained for another 50 epochs using a combination of the L1 and VGG~\citep{johnson2016perceptual} loss functions, where we scale the perceptual loss by a factor of 0.25. For both training phases, we set the learning rate to 1e-4 and use Adam~\citep{kingma2015adam} as optimizer with $\beta_1=0.5$ and $\beta_2=0.99$.

\section{Additional Results}
As a complement of Table~1 of the main paper, Table~\ref{tab:dresscode_categories} presents the complete quantitative results for each category of the Dress Code dataset. Our method, denoted as \ours, demonstrates superior performance compared to all competitors across all three Dress Code categories in terms of realism metrics such as FID and KID in both the paired and unpaired settings. 
When assessing input adherence metrics such as LPIPS and SSIM, our approach achieves better results than CP-VTON~\citep{wang2018toward} and CP-VTON$^{\text{\textdagger}}$ while still obtaining comparable results to PSAD~\citep{morelli2022dresscode}.

\begin{table}[t]
    \caption{Quantitative results per category on the \dataset dataset~\citep{morelli2022dresscode}.\vspace{-.15cm}}
    \label{tab:dresscode_categories}
    \setlength{\tabcolsep}{.35em}
    \resizebox{\linewidth}{!}{
    \begin{tabular}{lc cccccc}
    \toprule
     & & \multicolumn{6}{c}{\textbf{Upper-body}} \\
    \cmidrule{3-8}
    \textbf{Model} & & \textbf{LPIPS} $\downarrow$ & \textbf{SSIM} $\uparrow$ & \textbf{FID$_\text{p}$} $\downarrow$ & \textbf{KID$_\text{p}$} $\downarrow$ & \textbf{FID$_\text{u}$} $\downarrow$ & \textbf{KID$_\text{u}$} $\downarrow$ \\
    \midrule
    CP-VTON~\citep{wang2018toward} & & 0.176 & 0.851 & 46.47 & 33.82 & 48.31 & 35.25 \\
    CP-VTON$^\text{\textdagger}$~\citep{wang2018toward} & & 0.078 & 0.918 & 19.70 & 11.69 & 22.18 & 12.09 \\
    PSAD~\citep{morelli2022dresscode} & & 0.049 & 0.938 & 13.87 & 6.40 & 17.51 & 7.15  \\
    \rowcolor{LightCyan}
    \textbf{\ours} & & 0.049 & 0.928 & 9.53 & 1.98 & 13.26 & 2.67 \\
    \midrule
    & & \multicolumn{6}{c}{\textbf{Lower-body}} \\
    \cmidrule{3-8}
    \textbf{Model} & & \textbf{LPIPS} $\downarrow$ & \textbf{SSIM} $\uparrow$ & \textbf{FID$_\text{p}$} $\downarrow$ & \textbf{KID$_\text{p}$} $\downarrow$ & \textbf{FID$_\text{u}$} $\downarrow$ & \textbf{KID$_\text{u}$} $\downarrow$ \\
    \midrule
    CP-VTON~\citep{wang2018toward} & &  0.220 & 0.828 & 47.29 & 32.40 & 51.29 & 38.48 \\
    CP-VTON$^\text{\textdagger}$~\citep{wang2018toward} & & 0.083 & 0.913 & 18.85 & 10.33 & 18.85 & 10.24 \\
    PSAD~\citep{morelli2022dresscode} & & 0.051 & 0.932 & 13.14 & 5.59 & 19.68 & 8.90  \\
    \rowcolor{LightCyan}
    \textbf{\ours} & & 0.051 & 0.922 & 8.52 & 1.04 & 14.80 & 3.13 \\
    \midrule
    & & \multicolumn{6}{c}{\textbf{Dresses}} \\
    \cmidrule{3-8}
    \textbf{Model} & & \textbf{LPIPS} $\downarrow$ & \textbf{SSIM} $\uparrow$ & \textbf{FID$_\text{p}$} $\downarrow$ & \textbf{KID$_\text{p}$} $\downarrow$ & \textbf{FID$_\text{u}$} $\downarrow$ & \textbf{KID$_\text{u}$} $\downarrow$ \\
    \midrule
    CP-VTON~\citep{wang2018toward} & & 0.162 & 0.847 & 22.54 & 13.21 & 25.94 & 15.81 \\
    CP-VTON$^\text{\textdagger}$~\citep{wang2018toward} & & 0.123 & 0.863 & 18.75 & 11.07 & 21.83 & 12.31 \\
    PSAD~\citep{morelli2022dresscode} & & 0.074 & 0.885 & 12.38 & 4.68 & 17.07 & 6.66 \\
    \rowcolor{LightCyan}
    \textbf{\ours} & & 0.089 & 0.868 & 9.07 & 1.12 & 13.40 & 2.50 \\
    \bottomrule
    \end{tabular}
    }
\vspace{-.3cm}
\end{table}

\begin{figure*}[t]
    \centering
    \resizebox{.9\linewidth}{!}{
    \includegraphics[width=\linewidth]{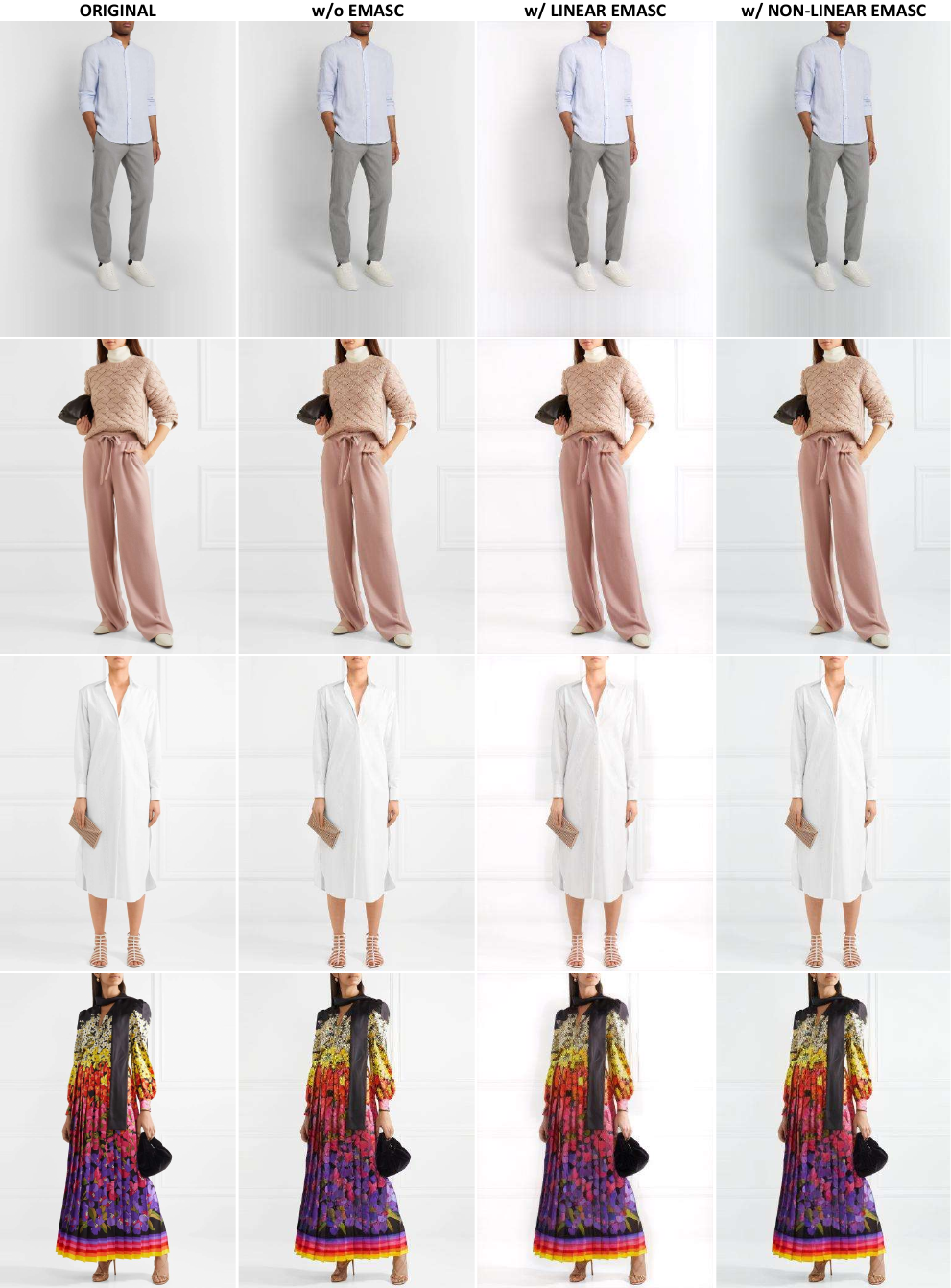}
    }
    \vspace{-0.2cm}
    \caption{Image reconstruction qualitative results from the autoencoder of Stable Diffusion on sample images from the Dress Code dataset. From left to right: the original image, the image from the out-of-the-box Stable Diffusion autoencoder without EMASC modules, the image from the autoencoder with linear EMASC connections, and the image from the autoencoder with non-linear EMASC connections.}
    \label{fig:vae_dresscode}
\end{figure*}

\begin{figure*}[t]
    \centering
    \resizebox{.9\linewidth}{!}{
    \includegraphics[width=\linewidth]{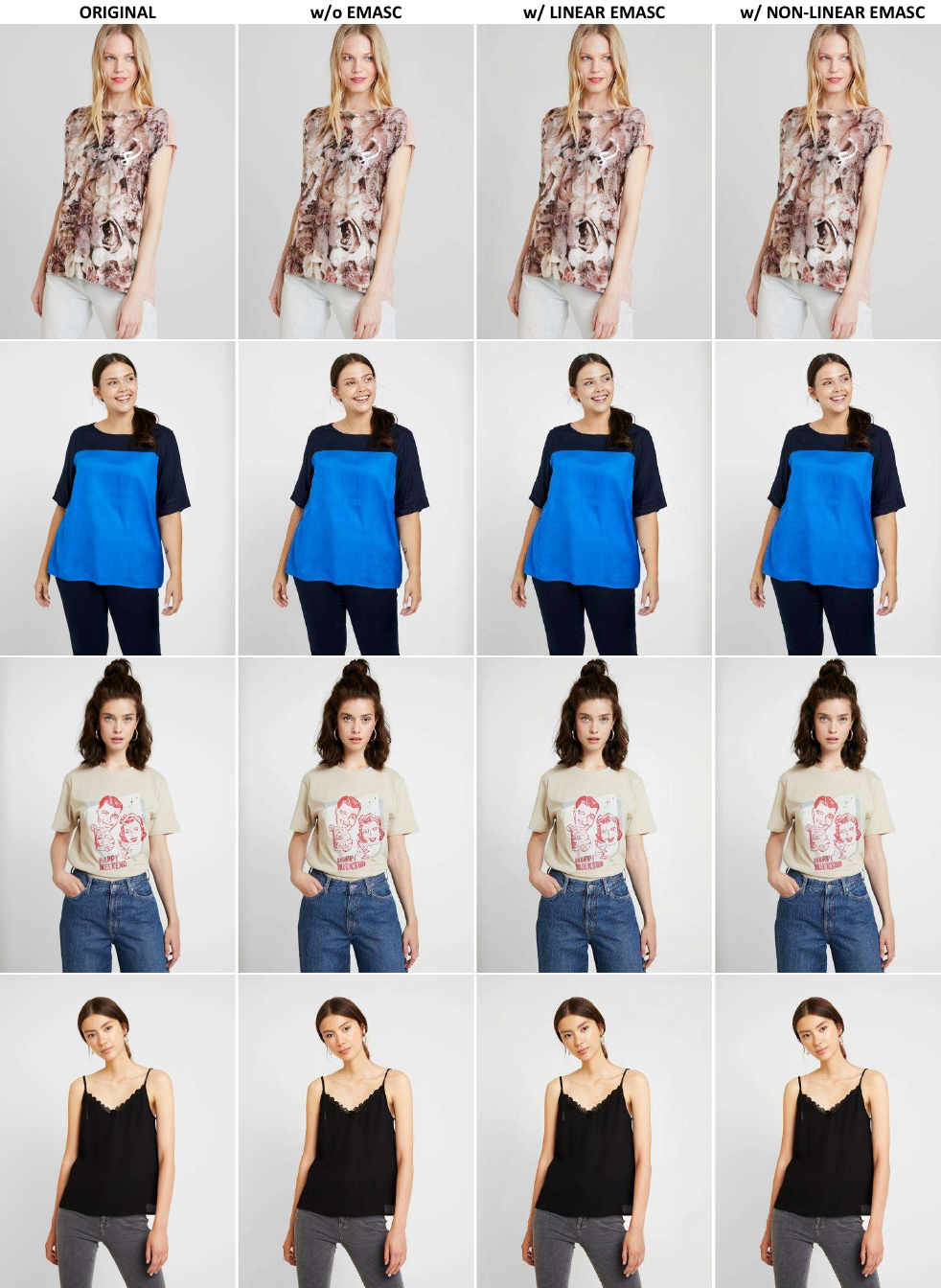}
    }
    \vspace{-0.2cm}
    \caption{Image reconstruction qualitative results from the autoencoder of Stable Diffusion on sample images from the VITON-HD dataset. From left to right: the original image, the image from the out-of-the-box Stable Diffusion autoencoder without EMASC modules, the image from the autoencoder with linear EMASC connections, and the image from the autoencoder with non-linear EMASC connections.}
    \label{fig:vae_vitonhd}
\end{figure*}

\tit{Qualitative results}
To provide further evidence of the effectiveness of the proposed EMASC modules in improving the quality of image reconstruction, we present additional qualitative results on the Dress Code and VITON-HD datasets. 
Specifically, Figures~\ref{fig:vae_dresscode} and \ref{fig:vae_vitonhd} depict examples of reconstructed images with and without the EMASC modules, showing the original image, the reconstructed image without EMASC, the reconstructed image with linear EMASC, and the reconstructed image with the proposed non-linear EMASC. The results show that the EMASC modules improve the quality of high-frequency human body details, such as hands, feet, and faces. In the Dress Code dataset (Figure~\ref{fig:vae_dresscode}), the EMASC module helps preserve the shapes of toes in the third and fourth rows. Similarly, in the VITON-HD dataset (Figure~\ref{fig:vae_vitonhd}), the EMASC module helps preserve the color and shape of eyes and avoid artifacts.

\begin{figure*}[t]
    \centering
    \resizebox{.94\linewidth}{!}{
    \includegraphics[width=\linewidth]{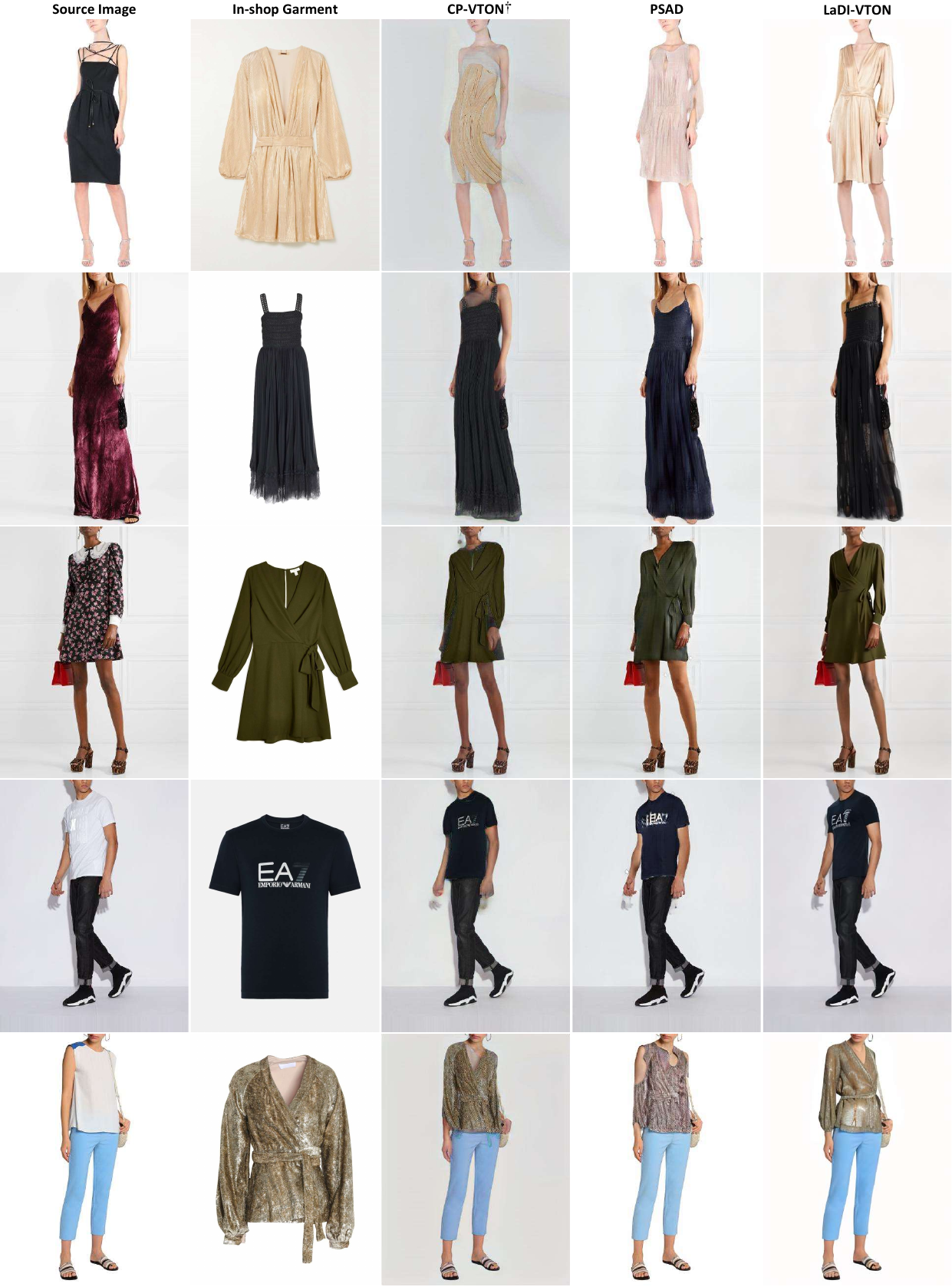}
    }
    \vspace{-0.2cm}
    \caption{Qualitative results generated by \ours and competitors on the Dress Code dataset. From left to right: the original image, the in-shop garment, and images generated by CP-VTON$^{\text{\textdagger}}$~\citep{wang2018toward}, PSAD~\citep{morelli2022dresscode}, \ours (ours).}
    \label{fig:supp_method_dresscode}
\end{figure*}

\begin{figure*}[t]
    \centering
    \resizebox{.94\linewidth}{!}{
    \includegraphics[width=\linewidth]{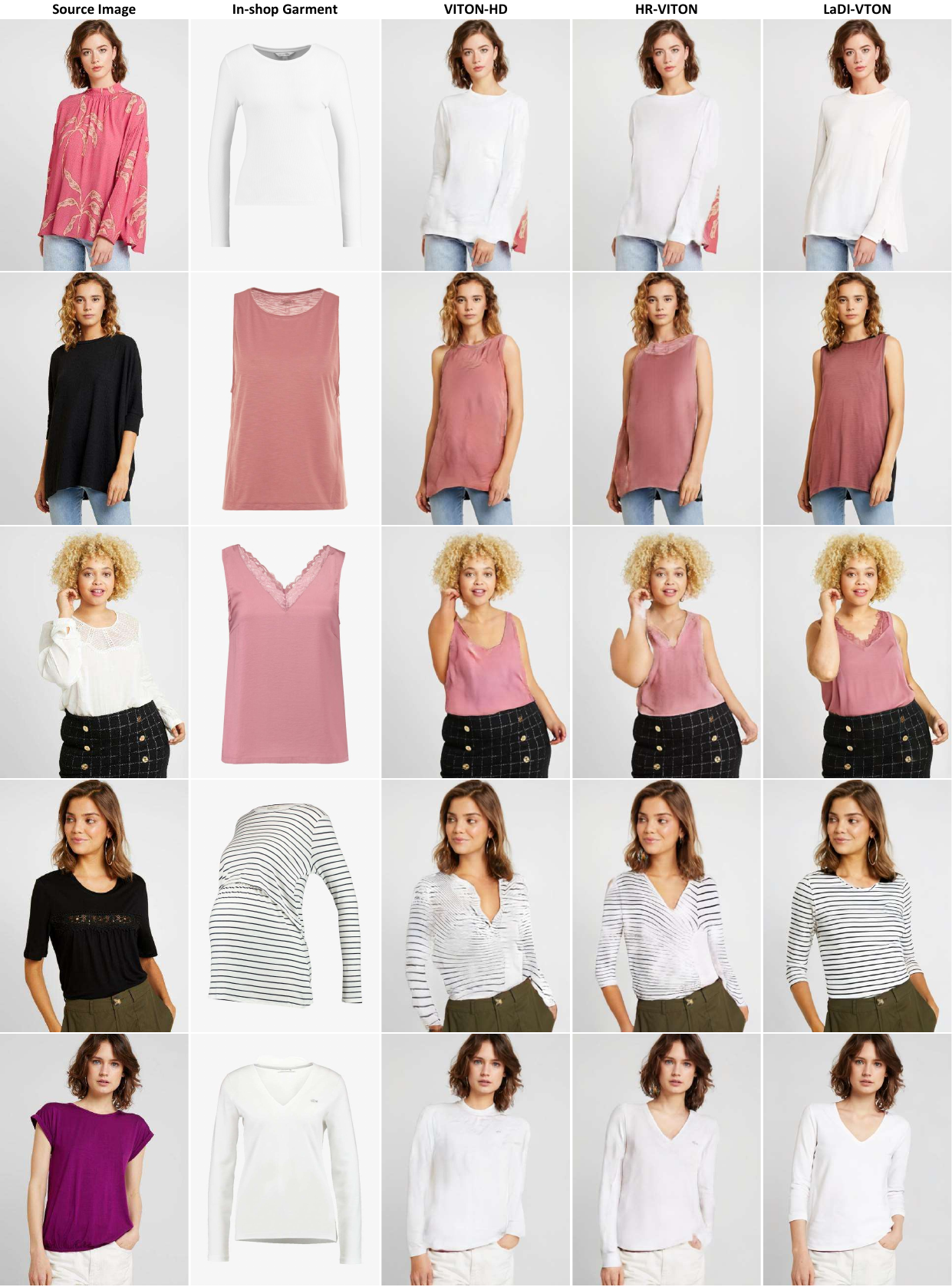}
    }
    \vspace{-0.2cm}
    \caption{Qualitative results generated by \ours and competitors on the Dress Code dataset. From left to right: the original image, the in-shop garment, and images generated by VITON-HD~\citep{choi2021viton}, HR-VITON~\citep{lee2022high}, \ours (ours).}
    \label{fig:supp_method_vitonhd}
\end{figure*}

Finally, we report additional qualitative results of our proposed virtual try-on pipeline and its competitors on two datasets. Our findings, as illustrated in Figure~\ref{fig:supp_method_dresscode} and Figure~\ref{fig:supp_method_vitonhd}, validate the quantitative results shown in the main paper, demonstrating that the \ours method produces more realistic outcomes. As it can be seen, our approach surpasses other methods in terms of visual quality, generating highly realistic images that showcase the effectiveness and robustness of our pipeline. These additional qualitative results further support the conclusion drawn from the quantitative analysis and have significant implications for the future development of virtual try-on technology.

\section{Limitations}
\begin{figure}[t]
    \centering
    \includegraphics[width=\linewidth]{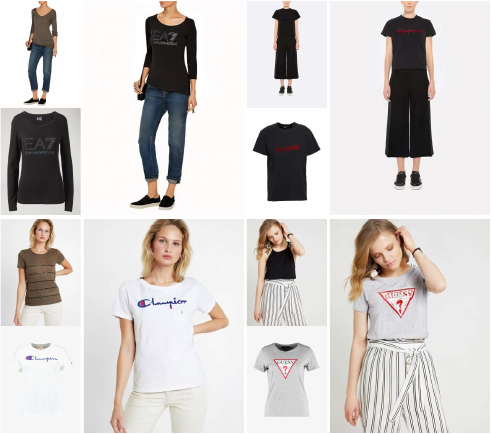}
    \caption{Failure cases of \ours on Dress Code~\citep{morelli2022dresscode} (1st row) and VITON-HD~\citep{choi2021viton} (2nd row)}
    \label{fig:failure_cases}
    \vspace{-.3cm}
\end{figure}

Beyond the difficulties in reproducing high-frequency details, current Stable Diffusion-based architectures still present some deficiencies in reproducing readable and coherent textual details. On this line, one of the key limitations of \ours is that it can not always synthesize logos and words depicted on the try-on garment faithfully. Some failure cases of the proposed approach are depicted in Figure \ref{fig:failure_cases}. Our model is able to well reproduce the general shape of logos and texts preserving the overall structure of the pattern (see second row), but however, struggles to define precise and highly comprehensible letters or numbers.
We argue that this flaw results from our model's reliance on Stable Diffusion, and it could be addressed by using a non-latent diffusion approach, however, with higher computational load and resource demand.

\balance 
\bibliographystyle{ACM-Reference-Format}
\bibliography{bibliography}

\end{document}